\newif\ifijcai\ijcaifalse 

\documentclass[10pt,twocolumn]{article}

\renewenvironment{abstract}{\centerline{\bf
Abstract}\vspace{0.5ex}\begin{quote}\small}{\par\end{quote}\vskip 1ex}
\newenvironment{keywords}{\centerline{\bf
Key Words}\vspace{0.5ex}\begin{quote}\small}{\par\end{quote}\vskip 1ex}

\def\toinfty#1{\stackrel{#1\to\infty}{\longrightarrow}}
\def\gtapprox{\buildrel{\lower.7ex\hbox{$>$}}\over
                       {\lower.7ex\hbox{$\sim$}}}
\def\nq{\hspace{-1em}}

\def\ignore#1{}

\def\hbar{h\!\!\!\!^{-}\,}

\def\beq{\begin{equation}}
\def\eeq{\end{equation}}
\def\beqn{\begin{displaymath}}
\def\eeqn{\end{displaymath}}
\def\bqa{\begin{equation}\begin{array}{c}}
\def\eqa{\end{array}\end{equation}}
\def\bqan{\begin{displaymath}\begin{array}{c}}
\def\eqan{\end{array}\end{displaymath}}
\def\pb{\underline}                       
\def\pb#1{\underline{#1}}                 
\def\maxarg{\mathop{\rm maxarg}}          
\def\hh#1{{\dot{#1}}}                     
\def\best{*}                              

\ifijcai\def\paragraph#1{{\bf #1}}\fi

\title{\bf \Large Towards a Universal Theory of Artificial Intelligence
based on \\ Algorithmic Probability and Sequential Decision Theory}

{\author{ Marcus Hutter \\[2mm]
  {\small IDSIA, Galleria 2, CH-6928 Manno-Lugano, Switzerland}  \\
  {\small marcus@idsia.ch \qquad http://www.idsia.ch} \\
  {\small Technical Report IDSIA-14-00, 16. December 2000}
}

\date{}

\begin{document}

\maketitle

\begin{abstract}
Decision theory formally solves the problem of rational agents in
uncertain worlds if the true environmental probability
distribution is known. Solomonoff's theory of universal induction
formally solves the problem of sequence prediction for unknown
distribution. We unify both theories and give strong arguments
that the resulting universal AI$\xi$ model behaves optimal in any
computable environment. The major drawback of the AI$\xi$ model is
that it is uncomputable. To overcome this problem, we construct a
modified algorithm AI$\xi^{tl}$, which is still superior to any
other time $t$ and space $l$ bounded agent. The computation time
of AI$\xi^{tl}$ is of the order $t\!\cdot\!2^l$.\\
\end{abstract}

\ifijcai\else
\fi


\section{Introduction}\label{int}

The most general framework for Artificial Intelligence is the
picture of an {\em agent} interacting with an environment
\cite{Russell:95}. If the goal is not pre-specified, the agent has
to learn by occasional reinforcement feedback \cite{Sutton:98}. If
the agent shall be universal, no assumption about the environment
may be made, besides that there {\it exists} some exploitable
structure at all. We may ask for the most intelligent way an agent
could behave, or, about the optimal way of learning in terms of
real world interaction cycles. {\em Decision theory}
formally\footnote{With a formal solution we mean a rigorous
mathematically definition, uniquely specifying the solution. For
problems considered here this always implies the existence of an
algorithm which asymptotically converges to the correct solution.}
solves this problem only if the true environmental probability
distribution is known (e.g. Backgammon)
\cite{Bellman:57,Bertsekas:96}. \cite{Solomonoff:64,Solomonoff:78}
formally solves the problem of {\em induction} if the true
distribution is unknown but only if the agent cannot influence the
environment (e.g.\ weather forecasts) \cite{Li:97}. We combine
both ideas and get {\em a parameterless model AI$\xi$ of an acting
agent which we claim to behave optimally in any computable
environment} (e.g.\ prisoner or auction problems, poker, car
driving). To get an effective solution, a modification
AI$\xi^{tl}$, superior to any other time $t$ and space $l$ bounded
agent, is constructed. The computation time of AI$\xi^{tl}$ is of
the order $t\!\cdot\!2^l$. The main goal of this work is to derive
and discuss the AI$\xi$ and the AI$\xi^{tl}$ model, and to clarify
the meaning of {\it universal}, {\it optimal}, {\it superior},
{\it etc}. Details can be found in \cite{Hutter:00f}.

\section{Rational Agents \& Sequential Decisions}\label{secAImurec}

\paragraph{Agents in probabilistic environments:}
A very general framework for intelligent systems is that of
rational agents \cite{Russell:95}. In cycle $k$, an agent performs
{\em action} $y_k\!\in\!Y$ (output word) which results in a {\em
perception} $x_k\!\in\!X$ (input word), followed by cycle
$k\!+\!1$ and so on. If agent and environment are deterministic
and computable, the entanglement of both can be modeled by two
Turing machines with two common tapes (and some private tapes)
containing the action stream $y_1y_2y_3...$ and the perception
stream $x_1x_2x_3...$ (The meaning of $x_k\!\equiv\!x'_kr_k$ is
explained in the next paragraph):

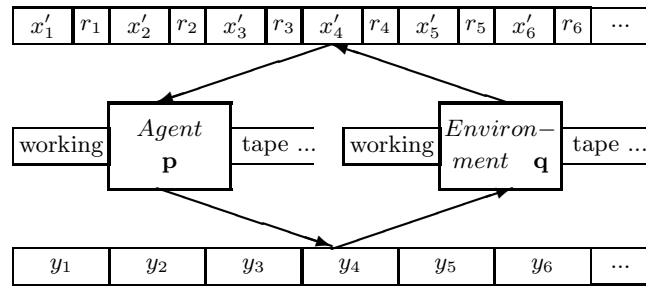
\begin{center}\label{cyberpic}
\small\unitlength=0.8mm
\special{em:linewidth 0.4pt}
\linethickness{0.4pt}
\begin{picture}(106,47)
\thinlines
\put(1,41){\framebox(10,6)[cc]{$x'_1$}}
\put(11,41){\framebox(6,6)[cc]{$r_1$}}
\put(17,41){\framebox(10,6)[cc]{$x'_2$}}
\put(27,41){\framebox(6,6)[cc]{$r_2$}}
\put(33,41){\framebox(10,6)[cc]{$x'_3$}}
\put(43,41){\framebox(6,6)[cc]{$r_3$}}
\put(49,41){\framebox(10,6)[cc]{$x'_4$}}
\put(59,41){\framebox(6,6)[cc]{$r_4$}}
\put(65,41){\framebox(10,6)[cc]{$x'_5$}}
\put(75,41){\framebox(6,6)[cc]{$r_5$}}
\put(81,41){\framebox(10,6)[cc]{$x'_6$}}
\put(91,41){\framebox(6,6)[cc]{$r_6$}}
\put(102,44){\makebox(0,0)[cc]{...}}
\put(1,1){\framebox(16,6)[cc]{$y_1$}}
\put(17,1){\framebox(16,6)[cc]{$y_2$}}
\put(33,1){\framebox(16,6)[cc]{$y_3$}}
\put(49,1){\framebox(16,6)[cc]{$y_4$}}
\put(65,1){\framebox(16,6)[cc]{$y_5$}}
\put(81,1){\framebox(16,6)[cc]{$y_6$}}
\put(102,4){\makebox(0,0)[cc]{...}}
\put(97,47){\line(1,0){9}}
\put(97,41){\line(1,0){9}}
\put(97,7){\line(1,0){9}}
\put(97,1){\line(0,0){0}}
\put(97,1){\line(1,0){9}}
\put(1,21){\framebox(16,6)[cc]{working}}
\thicklines
\put(17,17){\framebox(20,14)[cc]{$\displaystyle{Agent\atop\bf p}$}}
\thinlines
\put(37,27){\line(1,0){14}}
\put(37,21){\line(1,0){14}}
\put(39,24){\makebox(0,0)[lc]{tape ...}}
\put(56,21){\framebox(16,6)[cc]{working}}
\thicklines
\put(72,17){\framebox(20,14)[cc]{$\displaystyle{Environ-\atop ment\quad\bf q}$}}
\thinlines
\put(92,27){\line(1,0){14}}
\put(92,21){\line(1,0){14}}
\put(94,24){\makebox(0,0)[lc]{tape ...}}
\thicklines
\put(54,41){\vector(-3,-1){29}}
\put(84,31){\vector(-3,1){30}}
\put(54,7){\vector(3,1){30}}
\put(25,17){\vector(3,-1){29}}
\end{picture}
\end{center}

$p$ is the {\em policy} of the agent interacting with environment
$q$. We write $p(x_{<k})\!=\!y_{1:k}$ to denote the output
$y_{1:k}\!\equiv\!y_1...y_k$ of the agent $p$ on input
$x_{<k}\!\equiv\!x_1...x_{k-1}$ and similarly $q(y_{1:k})\!=\!x_{1:k}$
for the environment $q$. We call Turing machines
$p$ and $q$ behaving in this way {\it chronological}. In the more
general case of a {\em probabilistic environment}, given the
history $y\!x_{<k}y_k\!\equiv\!y_1x_1...y_{k-1}x_{k-1}y_k$, the
probability that the environment leads to perception $x_k$ in
cycle $k$ is (by definition) $\mu(y\!x_{<k}y\!\pb x_k)$. The
underlined argument $\pb x_k$ in $\mu$ is a probability variable
and the other non-underlined arguments $y\!x_{<k}y_k$ represent
conditions. We call probability distributions like $\mu$ {\it
chronological}.

\paragraph{The AI$\mu$ Model:}
The goal of the agent is to maximize future {\em rewards}, which are
provided by the environment through the inputs $x_k$. The inputs
$x_k\!\equiv\!x'_kr_k$ are divided into a regular part $x'_k$ and
some (possibly empty or delayed) reward $r_k$. The $\mu$-expected
reward sum of future cycles $k$ to $m$ with outputs
$y_{k:m}\!=\!y_{k:m}^p$ generated by the agent's policy $p$
can be written compactly as
\begin{equation}\label{vpdef}
  V_\mu^p(\hh y\!\hh x_{<k}) \!:=\!\!
  \!\!\sum_{x_k...x_m}\!\!
  (r_k\!+...+\!r_m)
  \mu(\hh y\!\hh x_{<k}y\!\pb x_{k:m}),
\end{equation}
where $m$ is the {\em lifespan} of the agent,
and the dots above $\hh y\!\hh x_{<k}$
indicate the actual action and perception history.
The $\mu$-expected reward sum of future cycles $k$ to $m$
with outputs $y_i$ generated by the {\em ideal agent}, which
maximizes the expected future rewards is
\begin{equation}\label{voptdef}
  V_\mu^\best(\hh y\!\hh x_{<k}) :=
  \max_{y_k}\!\sum_{x_k}...
  \max_{y_{m}}\!\sum_{x_{m}}
  (r_k\!+...+\!r_m)
  \mu(\hh y\!\hh x_{<k}y\!\pb x_{k:m}),
\end{equation}
i.e.\ the best expected credit
is obtained by averaging over the $x_i$ and
maximizing over the $y_i$. This has to be done in chronological
order to correctly incorporate the dependency of $x_i$ and $y_i$
on the history. The output $\hh y_k$, which achieves the maximal value
defines {\em the AI$\mu$ model}:
\begin{equation}\label{ydotrec}
  \hh y_k :=
  \maxarg_{y_k}\!\sum_{x_k}...
  \max_{y_{m}}\!\sum_{x_{m}}
  (r_k\!+...+\!r_m)
  \mu(\hh y\!\hh x_{<k}y\!\pb x_{k:m}).
\end{equation}
The AI$\mu$ model is optimal in the sense that no other policy
leads to higher $\mu$-expected reward. A detailed derivation and
other recursive and functional versions can be found in
\cite{Hutter:00f}.

\paragraph{Sequential decision theory:}
Eq.\ (\ref{ydotrec}) is essentially an Expectimax algorithm/sequence.
One can relate (\ref{ydotrec}) to the Bellman equations
\cite{Bellman:57} of sequential decision theory by identifying
complete histories $y\!x_{<k}$ with states, $\mu(y\!x_{<k}y\!\pb
x_k)$ with the state transition matrix, $V_\mu^\best(y\!x_{<k})$
with the value of history/state $y\!x_{<k}$, and $y_k$ with the
action in cycle $k$ \cite{Russell:95,Hutter:00f}.
Due to the use of complete histories as state space, the AI$\mu$
model neither assumes stationarity, nor the Markov property, nor
complete accessibility of the environment. Every state occurs at
most once in the lifetime of the system.
As we have in mind a universal system with complex interactions,
the action and perception spaces $Y$ and $X$ are huge (e.g.\ video
images), and every action or perception itself occurs usually only
once in the lifespan $m$ of the agent. As there is no (obvious)
universal similarity relation on the state space, an effective
reduction of its size is impossible, but there is no principle
problem in determining $\hh y_k$ as long as $\mu$ is known and
computable and $X$, $Y$ and $m$ are finite.

\paragraph{Reinforcement learning:}
Things dramatically change if $\mu$ is unknown. Reinforcement
learning algorithms \cite{Kaelbling:96,Sutton:98,Bertsekas:96} are
commonly used in this case to learn the unknown $\mu$. They
succeed if the state space is either small or has effectively been
made small by generalization or function approximation techniques.
In any case, the solutions are either ad hoc, work in restricted
domains only, have serious problems with state space exploration
versus exploitation, or have non-optimal learning rate. There is
no universal and optimal solution to this problem so far. In the
Section \ref{secAIxi} we present a new model and argue that it
formally solves all these problems in an optimal way. The true
probability distribution $\mu$ will not be learned directly, but
will be replaced by a universal prior $\xi$, which is shown to
converge to $\mu$ in a sense.

\section{Algorithmic Complexity and Universal Induction}\label{secAIsp}

\paragraph{The problem of the unknown environment:}
We have argued that currently there is no universal and optimal solution to
solving reinforcement learning problems. On the other hand,
\cite{Solomonoff:64} defined a universal scheme of inductive
inference, based on Epicurus' principle of multiple explanations,
Ockham's razor, and Bayes' rule for
conditional probabilities. For an excellent introduction
one should consult the book of \cite{Li:97}. In
the following we outline the theory and the basic results.

\paragraph{Kolmogorov complexity and universal probability:}
Let us choose some universal prefix Turing machine $U$ with
unidirectional binary input and output tapes and a bidirectional
working tape. We can then define the (conditional) prefix
Kolmogorov complexity
\cite{Chaitin:75,Gacs:74,Kolmogorov:65,Levin:74} as the
length $l$ of the shortest
program $p$, for which $U$ outputs the binary string
$x\!=\!x_{1:n}$ with $x_i\in\!\{0,1\}$:
$$
  K(x) \;:=\; \min_p\{l(p): U(p)=x\},
$$
and given $y$
$$
  K(x|y) \;:=\; \min_p\{l(p): U(p,y)=x\}.
$$
The {\em universal semimeasure} $\xi(\pb x)$ is defined as the
probability that the output of $U$
starts with $x$ when provided with fair coin flips on the input
tape \cite{Solomonoff:64,Solomonoff:78}. It is easy to see that
this is equivalent to the formal definition
\beq\label{xidef}
  \xi(\pb x)\;:=\;\sum_{p\;:\;\exists\omega:U(p)=x\omega}\nq 2^{-l(p)}
\eeq
where the sum is over minimal programs $p$ for which $U$
outputs a string starting with $x$. $U$ might be non-terminating.
As the short programs dominate the sum, $\xi$ is closely related
to $K(x)$ as $\xi(\pb x)=2^{-K(x)+O(K(l(x))}$. $\xi$ has the
important universality property \cite{Solomonoff:64} that it
dominates every computable probability
distribution $\rho$ up to a multiplicative factor depending only
on $\rho$ but not on $x$:
\beq\label{uni}
  \xi(\pb x) \;\geq\; 2^{-K(\rho)-O(1)}\!\cdot\!\rho(\pb x).
\eeq
The Kolmogorov complexity of a function like $\rho$ is defined as
the length of the shortest self-delimiting coding of a Turing
machine computing this function.
$\xi$ itself is {\it not} a probability
distribution\footnote{It is possible to normalize $\xi$ to a
probability distribution as has been done in
\cite{Solomonoff:78,Hutter:99} by giving up the enumerability of $\xi$.
Bounds (\ref{eukdist}) and (\ref{spebound}) hold for both
definitions.}.
We have $\xi(\pb{x0})\!+\!\xi(\pb{x1})\!<\!\xi(\pb
x)$ because there are programs $p$, which output just $x$, neither
followed by $0$ nor $1$. They just stop after printing $x$ or
continue forever without any further output. We will call a
function $\rho\!\geq 0$ with the properties
$\rho(\epsilon)\!\leq\!1$ and $\sum_{x_n}\rho(\pb
x_{1:n})\!\leq\!\rho(\pb x_{<n})$ a {\it semimeasure}. $\xi$ is a
semimeasure and (\ref{uni}) actually holds for all enumerable
semimeasures $\rho$.

\paragraph{Universal sequence prediction:}
(Binary) sequence prediction algorithms try to predict the
continuation $x_n$ of a given sequence $x_1...x_{n-1}$. In the
following we will assume that the sequences are drawn from
a probability distribution and that the true probability of a
string starting with $x_1...x_n$ is $\mu(\pb x_{1:n})$. The
probability of $x_n$ given $x_{<n}$ hence is $\mu(x_{<n}\pb x_n)$.
If we measure prediction quality as the number of correct
predictions, the best possible system predicts the $x_n$ with the
highest probability. Usually $\mu$ is unknown and the system can
only have some belief $\rho$ about the true distribution $\mu$.
Now the universal probability $\xi$
comes into play: \cite{Solomonoff:78} has proved
that the mean squared difference
between $\xi$ and $\mu$ is finite for computable $\mu$:
\beq\label{eukdist}
  \sum_{k=1}^\infty\sum_{x_{1:k}}\mu(\pb x_{<k})
  (\xi(x_{<k}\pb x_k)-\mu(x_{<k}\pb x_k))^2
\eeq
$$
  <\; \ln 2\!\cdot\!K(\mu)+O(1).
$$
A simplified proof can be found in \cite{Hutter:99}. So the
difference between $\xi(x_{<n}\pb x_n)$ and $\mu(x_{<n}\pb x_n)$ tends
to zero with $\mu$ probability $1$ for {\it any} computable
probability distribution $\mu$. The reason for the astonishing
property of a single (universal) function to converge to {\it any}
computable probability distribution lies in the fact that the set
of $\mu$-random sequences differ for different $\mu$. The
universality property (\ref{uni}) is the central ingredient for
proving (\ref{eukdist}).

\paragraph{Error bounds:}
Let SP$\rho$ be a probabilistic
sequence predictor, predicting $x_n$ with probability
$\rho(x_{<n}\pb x_n)$. If $\rho$ is only a semimeasure the
SP$\rho$ system might refuse any output in some cycles $n$.
Further, we define a deterministic sequence predictor
SP$\Theta_\rho$ predicting the $x_n$ with highest $\rho$
probability. $\Theta_\rho(x_{<n}\pb x_n)\!:=\!1$ for one $x_n$
with $\rho(x_{<n}\pb x_n)\!\geq\!\rho(x_{<n}\pb x'_n)\,\forall
x'_n$ and $\Theta_\rho(x_{<n}\pb x_n)\!:=\!0$ otherwise.
SP$\Theta_\mu$ is the best prediction scheme when $\mu$ is known.
If $\rho(x_{<n}\pb x_n)$ converges quickly to $\mu(x_{<n}\pb x_n)$ the
number of additional prediction errors introduced by using
$\Theta_\rho$ instead of $\Theta_\mu$ for prediction should be
small in some sense.
Let us define the total number of expected erroneous predictions
the SP$\rho$ system makes for the first $n$ bits:
\beq\label{esp}
  E_{n\rho} \;:=\; \sum_{k=1}^n\sum_{x_{1:k}}\mu(\pb x_{1:k})
  (1\!-\!\rho(x_{<k}\pb x_k)).
\eeq
The SP$\Theta_\mu$ system is best in the sense that
$E_{n\Theta_\mu}\!\leq\!E_{n\rho}$
for any $\rho$. In \cite{Hutter:99} it has been shown that
SP$\Theta_\xi$ is not much worse
\beq\label{spebound}
  E_{n\Theta_\xi}\!-\!E_{n\rho} \;\leq\;
  H+\sqrt{4E_{n\rho}H+H^2} \;=\;
  O(\sqrt{E_{n\rho}})
\eeq
$$
  \mbox{with}\quad H\;<\;\ln 2\!\cdot\!K(\mu)+O(1)
$$
and the tightest bound for $\rho\!=\!\Theta_\mu$. For finite
$E_{\infty\Theta_\mu}$, $E_{\infty\Theta_\xi}$ is finite too. For
infinite $E_{\infty\Theta_\mu}$,
$E_{n\Theta_\xi}/E_{n\Theta_\mu}\toinfty{n}1$ with rapid
convergence. One can hardly imagine any better prediction
algorithm as SP$\Theta_\xi$ without extra knowledge about the
environment. In \cite{Hutter:00e}, (\ref{eukdist}) and
(\ref{spebound}) have been generalized from binary to arbitrary
alphabet and to general loss functions. Apart from computational
aspects, which are of course very important, the problem of
sequence prediction could be viewed as essentially solved.

\section{The Universal AI$\xi$ Model}\label{secAIxi}

\paragraph{Definition of the AI$\xi$ Model:}
We have developed enough formalism to suggest our universal
AI$\xi$ model. All we have to do is to suitably generalize the
universal semimeasure $\xi$ from the last section and to replace
the true but unknown probability $\mu$ in the AI$\mu$ model by
this generalized $\xi$. In what sense this AI$\xi$ model is
universal and optimal will be discussed thereafter.

We define the generalized universal probability $\xi^{AI}$ as the
$2^{-l(q)}$ weighted sum over all chronological programs
(environments) $q$ which output $x_{1:k}$, similar to
(\ref{xidef}) but with $y_{1:k}$ provided on the ''input''
tape:
\beq\label{uniMAI}
  \xi(y\!\pb x_{1:k}) \;:=\;
  \nq\sum_{q:q(y_{1:k})=x_{1:k}}\nq 2^{-l(q)}.
\eeq
Replacing $\mu$ by $\xi$ in (\ref{ydotrec}) the
iterative AI$\xi$ system outputs
\beq\label{ydotxi}
  \hh y_k :=
  \maxarg_{y_k}\!\sum_{x_k}...
  \max_{y_m}\!\sum_{x_m}
  (c_k\!+...+\!c_m)
  \xi(\hh y\!\hh x_{<k}y\!\pb x_{k:m}).
\eeq
in cycle $k$ given the history $\hh y\!\hh x_{<k}$.

\paragraph{(Non)parameters of AI$\xi$:}
The AI$\xi$ model and its behaviour is completely defined by
(\ref{uniMAI}) and (\ref{ydotxi}). It (slightly) depends on the
choice of the universal Turing machine. The AI$\xi$ model also
depends on the choice of $X$ and $Y$, but we do
not expect any bias when the spaces are chosen sufficiently large
and simple, e.g. all strings of length $2^{16}$. Choosing $I\!\!N$
as word space would be ideal, but whether the maxima (or suprema)
exist in this case, has to be shown beforehand. The only
non-trivial dependence is on the horizon $m$. Ideally we would
like to chose $m\!=\!\infty$, but there are several subtleties
\ifijcai{discussed in \cite{Hutter:00f},}
\else{to be discussed later,}
\fi
which prevent at least a naive limit
$m\!\to\!\infty$. So apart from $m$ and unimportant details, the
AI$\xi$ system is uniquely defined by (\ref{ydotxi}) and
(\ref{uniMAI}) without adjustable parameters. It does not depend on
any assumption about the environment apart from being generated by
some computable (but unknown!) probability distribution as we will see.

\ifijcai\else
\paragraph{$\xi$ is only a semimeasure:}
One subtlety should be mentioned.
Like in the SP case, $\xi$ is
not a probability distribution but still satisfies the weaker
inequalities
\beq\label{chrf}
  \sum_{x_n}\xi(y\!\pb x_{1:n}) \;\leq\; \xi(y\!\pb x_{<n})
  \quad,\quad
  \xi(\epsilon) \;\leq\; 1
\eeq
Note, that the sum on the l.h.s.\ is {\it not} independent of
$y_n$ unlike for the chronological probability distribution $\mu$.
Nevertheless, it is bounded by something (the r.h.s) which is
independent of $y_n$. The reason is that the sum in (\ref{uniMAI})
runs over (partial recursive) chronological functions only and the
functions $q$ which satisfy $q(y_{1:n})=x_{<n}x'_n$ for some
$x'_n\!\in\!X$ are a subset of the functions satisfying
$q(y_{<n})=x_{<n}$. We will in general call functions satisfying
(\ref{chrf}) {\it chronological semimeasures}. The important point
is that the conditional probabilities (\ref{uniMAI}) are $\leq\!1$
like for true probability distributions.
\fi 

\paragraph{Universality of $\xi^{AI}$:}
It can be shown that $\xi^{AI}$ defined in
(\ref{uniMAI}) is universal and converges to $\mu^{AI}$
analogously to the SP case (\ref{uni}) and (\ref{eukdist}). The
proofs are generalizations from the SP case. The actions $y$ are pure
spectators and cause no difficulties in the generalization. This
will change when we analyze error/value bounds analogously to
(\ref{spebound}). The major difference when incorporating $y$ is
that in (\ref{uni}), $U(p)=x\omega$ produces strings starting with $x$,
whereas in (\ref{uniMAI}) we can demand $q$ to output exactly $n$
words $x_{1:n}$ as $q$ knows $n$ from the number of input words
$y_1...y_n$.
$\xi^{AI}$ dominates all {\em chronological enumerable
semimeasures}
\beq\label{uniaixi}
  \xi(y\!\pb x_{1:n}) \;\geq\;
  2^{-K(\rho)-O(1)}\rho(y\!\pb x_{1:n}).
\eeq
$\xi$ is a universal element in the sense of (\ref{uniaixi})
in the set of all enumerable chronological semimeasures. This can
be proved even for infinite (countable) alphabet
\cite{Hutter:00f}.

\paragraph{Convergence of $\xi^{AI}$ to $\mu^{AI}$:}
From (\ref{uniaixi}) one can show
$$
  \sum_{k=1}^n\sum_{x_{1:k}}\mu(y\!\pb x_{<k})
  \Big(\mu(y\!x_{<k}y\!\pb x_k)-\xi(y\!x_{<k}y\!\pb x_k)\Big)^2
$$
\beq\label{eukdistxi}
  \;<\; \ln 2\!\cdot\!K(\mu)+O(1)
\eeq
for computable chronological measures $\mu$. The main
complication in generalizing (\ref{eukdist}) to (\ref{eukdistxi})
is the generalization to non-binary alphabet \cite{Hutter:00e}.
The $y$ are, again, pure spectators.
(\ref{eukdistxi}) shows that the $\mu$-expected
squared difference of $\mu$ and $\xi$ is finite for computable
$\mu$. This, in turn, shows that $\xi(y\!x_{<k}y\!\pb x_k)$
converges to $\mu(y\!x_{<k}y\!\pb x_k)$ for $k\!\to\!\infty$ with $\mu$
probability 1. If we take a finite product of $\xi'$s and use
Bayes' rule, we see that also $\xi(y\!x_{<k}y\!\pb x_{k:k+r})$
converges to $\mu(y\!x_{<k}y\!\pb x_{k:k+r})$. More generally, in case of
a bounded horizon $h_k\equiv m_k\!-\!k\!+\!1 \leq h_{max}\!<\!\infty$, it follows that
\beq\label{aixitomu}
  \xi(y\!x_{<k}y\!\pb x_{k:m_k}) \toinfty{k} \mu(y\!x_{<k}y\!\pb x_{k:m_k})
\eeq
Convergence is only guaranteed for one (e.g.\ the true) i/o
sequence $\hh y\!\hh x_{<k}\hh y\!\hh x_{k:m_k}$ but not for
alternate sequences $\hh y\!\hh x_{<k}y\!x_{k:m_k}$. Since
(\ref{ydotxi}) takes an average over all possible future actions
and perceptions $y\!x_{k:m_k}$; not only the one which will
finally occur, (\ref{aixitomu}) does not guarantee $\hh y_k^\xi\!\to\!\hh
y_k^\mu$. This
gap is already present in the SP$\Theta_\rho$ models, but
nevertheless good error bounds could be proved. This gives
confidence that the outputs $\hh y_k$ of the AI$\xi$ model
(\ref{ydotxi}) could converge to the outputs $\hh y_k$ of the
AI$\mu$ model (\ref{ydotrec}), at least for a bounded horizon
$h_k$. The problems with a fixed horizon $m_k\!=\!m$ and especially
$m\!\to\!\infty$
\ifijcai{are discussed in \cite{Hutter:00f}.}
\else{will be discussed later.}
\fi

\paragraph{Universally optimal AI systems:}
We want to call an AI model {\it universal}, if it is
$\mu$-independent (unbiased, model-free) and is able to solve any
solvable problem and learn any learnable task. Further, we call a
universal model, {\it universally optimal}, if there is no
program, which can solve or learn significantly faster (in terms
of interaction cycles). As the AI$\xi$ model is parameterless,
$\xi$ converges to $\mu$ in the sense of
(\ref{eukdistxi},\ref{aixitomu}), the AI$\mu$ model is itself
optimal, and we expect no other model to converge faster to
AI$\mu$ by analogy to SP (\ref{spebound}),
\beqn
  \mbox{\it we expect AI$\xi$ to be universally optimal.}
\eeqn
This is our main claim. Further support is given in
\cite{Hutter:00f} by a detailed analysis of the behaviour of
AI$\xi$ for various problem classes, including prediction,
optimization, games, and supervised learning.

\ifijcai\else
\paragraph{The choice of the horizon:}
The only significant arbitrariness in the AI$\xi$ model lies in
the choice of the lifespan $m$ or the
$h_k\!\equiv\!m_k\!-\!k\!+\!1$ if we allow a cycle dependent $m$.
We will not discuss ad hoc choices of $h_k$ for specific problems.
We are interested in universal choices. The book of
\cite{Bertsekas:95b} thoroughly discusses the mathematical
problems regarding infinite horizon systems.

In many cases the time we are willing to run a system depends on
the quality of its actions. Hence, the lifetime, if finite at all,
is not known in advance. Exponential discounting
$r_k\!\to\!r_k\!\cdot\!\gamma^k$ solves the mathematical problem
of $m\!\to\!\infty$ but is no real solution, since an effective
horizon $h\sim\ln{1\over\gamma}$ has been introduced. The scale
invariant discounting $r_k\!\to\!r_k\!\cdot\!k^{-\alpha}$ has a
dynamic horizon $h\sim\!k$. This choice has some appeal, as it
seems that humans of age $k$ years usually do not plan their lives
for more than the next $\sim k$ years. From a practical point of
view this model might serve all needs, but from a theoretical
point we feel uncomfortable with such a limitation in the horizon
from the very beginning. A possible way of taking the limit
$m\!\to\!\infty$ without discounting and its problems can be found
in \cite{Hutter:00f}.

Another objection against too large choices of $m_k$
is that $\xi(y\!x_{<k}y\!\pb x_{k:m_k})$ has been proved to be a
good approximation of $\mu(y\!x_{<k}y\!\pb x_{k:m_k})$ only for
$k\!\gg\!h_k$, which is never satisfied for
$m_k\!=\!m\!\to\!\infty$.
On the other hand it may turn out that the rewards
$r_{k'}$ for $k'\!\gg\!k$, where $\xi$ may no longer be trusted as
a good approximation of $\mu$, are in a sense randomly
disturbed with decreasing influence on the choice of $\hh y_k$.
This claim is supported by the forgetfulness property of $\xi$
\ifijcai\else{(see next section)}\fi
and can be proved when restricting to
factorizable environments \cite{Hutter:00f}.

We are not sure whether the choice of $m_k$ is of marginal
importance, as long as $m_k$ is chosen sufficiently large and of
low complexity, $m_k=2^{2^{16}}$ for instance, or whether the
choice of $m_k$ will turn out to be a central topic for the
AI$\xi$ model or for the planning aspect of any universal AI
system in general. Most if not all problems in agent design of
balancing exploration and exploitation vanish by a sufficiently
large choice of the (effective) horizon and/or a sufficiently
general prior. We suppose that the limit $m_k\!\to\!\infty$ for
the AI$\xi$ model results in correct behaviour for weakly
separable (defined in the next section) $\mu$, and that even the
naive limit $m\!\to\!\infty$ may exist.
\fi

\paragraph{Value bounds and separability concepts:}
The values  $V_\rho^\best$ associated with the AI$\rho$ systems
correspond roughly to the negative error measure $-E_{n\rho}$ of
the SP$\rho$ systems. In the SP case we were interested in small
bounds for the error excess $E_{n\Theta_\xi}\!-\!E_{n\rho}$.
Unfortunately, simple value bounds for AI$\xi$ or any other AI system in terms of
$V^\best$ analogously to the error bound (\ref{spebound}) can not
hold \cite{Hutter:00f}. We even have difficulties in specifying
what we can expect to hold for AI$\xi$ or any AI system which
claims to be universally optimal. In SP, the only important
property of $\mu$ for proving error bounds was its complexity
$K(\mu)$. In the AI case, there are no useful bounds in terms of
$K(\mu)$ only. We either have to study restricted problem classes
or consider bounds depending on other properties of $\mu$, rather
than on its complexity only. In \cite{Hutter:00f} the difficulties
are exhibited by two examples. Several concepts, which might be
useful for proving value bounds are introduced and discussed. They
include forgetful, relevant, asymptotically learnable, farsighted,
uniform, (generalized) Markovian, factorizable and (pseudo)
passive $\mu$. They are approximately sorted in the order of
decreasing generality and are called {\it separability concepts}.
A first weak bound for passive $\mu$ is proved.

\section{Time Bounds and Effectiveness}\label{secTime}

\paragraph{Non-effectiveness of AI$\xi$:}
$\xi$ is not a computable but
only an enumerable semimeasure. Hence, the output $\hh y_k$ of the
AI$\xi$ model is only asymptotically computable. AI$\xi$ yields an
algorithm that produces a sequence of trial outputs eventually
converging to the correct output $\hh y_k$, but one can never be sure
whether one has already reached it. Besides this, convergence
is extremely slow, so this type of asymptotic computability is of
no direct (practical) use. Furthermore, the replacement
of $\xi$ by time-limited versions \cite{Li:91,Li:97}, which is
suitable for sequence prediction, has been shown to fail for the
AI$\xi$ model \cite{Hutter:00f}.
This leads to the issues addressed next.

\paragraph{Time bounds and effectiveness:}
Let $\tilde p$ be a policy which calculates an acceptable output
within a reasonable time $\tilde t$ per cycle. This sort of
computability assumption, namely, that a general purpose computer
of sufficient power and appropriate program is able to behave in
an intelligent way, is the very basis of AI research. Here it is
not necessary to discuss what exactly is meant by
'reasonable time/intelligence' and 'sufficient power'. What we are
interested in is whether there is a computable version
AI$\xi^{\tilde t}$ of the AI$\xi$ system which is superior or
equal to any program $p$ with computation time per cycle of at
most $\tilde t$.

What one can realistically hope to construct is an AI$\xi^{\tilde
t\tilde l}$ system of computation time $c\!\cdot\!\tilde t$ per
cycle for some constant $c$. The idea is to run all programs $p$
of length $\leq\!\tilde l\!:=\!l(\tilde p)$ and time $\leq\!\tilde
t$ per cycle and pick the best output in the sense of maximizing
the {\em universal value} $V_\xi^\best$. The total computation time is
$c\!\cdot\!\tilde t$ with $c\!\approx\!2^{\tilde l}$. Unfortunately
$V_\xi^\best$ can not be used directly since this measure is also
only semi-computable and the approximation quality by using
computable versions of $\xi$ given a time of order
$c\!\cdot\!\tilde t$ is crude \cite{Li:97,Hutter:00f}. On the
other hand, we {\it have} to use a measure which converges
$V_\xi^\best$ for $\tilde t,\tilde l\!\to\!\infty$, since the
AI$\xi^{\tilde t\tilde l}$ model should converge to the AI$\xi$ model
in that case.

\paragraph{Valid approximations:}
A solution satisfying the above conditions is suggested in
\cite{Hutter:00f}. The main idea is to consider {\em extended
chronological incremental policies} $p$, which in addition to the
regular output $y_k^p$ {\em rate} their own output with $w_k^p$. The
AI$\xi^{\tilde t\tilde l}$ model selects the output $\hh y_k\!=\!y_k^p$
of the policy $p$ with highest rating $w_k^p$. $p$ might suggest
any output $y_k^p$ but it is not allowed to rate itself with an
arbitrarily high $w_k^p$ if one wants $w_k^p$ to be a reliable
criterion for selecting the best $p$. One must demand that no
policy $p$ is allowed to claim that it is better than it actually
is. In \cite{Hutter:00f} a (logical) predicate VA($p$), called
{\it valid approximation}, is defined, which is true if, and only
if, $p$ {\it always} satisfies $w_k^p\!\leq\!V_\xi^p(y\!x_{<k})$, i.e. never
overrates itself. $V_\xi^p(y\!x_{<k})$ is the $\xi$ expected
future reward under policy $p$. Valid policies $p$ can then be
(partially) ordered w.r.t.\ their rating $w_k^p$.

\paragraph{The universal time bounded AI$\xi^{\tilde t\tilde l}$ system:}
In the following, we describe the algorithm $p^\best$ underlying
the universal time bounded AI$\xi^{\tilde t\tilde l}$ system. It
is essentially based on the selection of the best algorithms
$p_k^\best$ out of the time ${\tilde t}$ and length ${\tilde l}$
bounded policies $p$, for which there exists a proof $P$ of
VA($p$) with length $\leq\!l_P$.

\begin{enumerate}\parskip=0ex\parsep=0ex\itemsep=0ex
\item Create all binary strings of length $l_P$ and interpret each
as a coding of a mathematical proof in the same formal logic system in
which VA($\cdot$) has been formulated. Take those strings
which are proofs of VA($p$) for some $p$ and keep the
corresponding programs $p$.
\item Eliminate all $p$ of length $>\!\tilde l$.
\item Modify all $p$ in the following way: all output $w_k^py_k^p$
is temporarily written on an auxiliary tape. If $p$ stops in $\tilde t$
steps the internal 'output' is copied to the output tape. If $p$
does not stop after $\tilde t$ steps a stop is forced and $w_k^p\!=\!0$
and some arbitrary $y_k^p$ is written on the output tape. Let ${\cal P}$ be
the set of all those modified programs.
\item Start first cycle: $k\!:=\!1$.
\item\label{pbestloop} Run every $p\!\in\!{\cal P}$ on extended input
$\hh y\!\hh x_{<k}$, where all outputs are redirected to some auxiliary
tape:
$p(\hh y\!\hh x_{<k})\!=\!w_1^py_1^p...w_k^py_k^p$. This step is
performed incrementally by adding $\hh y\!\hh x_{k-1}$ for $k\!>\!1$ to
the input tape and continuing the computation of the previous
cycle.
\item Select the program $p$ with highest rating $w_k^p$:
$p_k^\best\!:=\!\maxarg_pw_k^p$.
\item Write $\hh y_k\!:=\!y_k^{p_k^\best}$ to the output tape.
\item Receive input $\hh x_k$ from the environment.
\item Begin next cycle: $k\!:=\!k\!+\!1$, goto step
\ref{pbestloop}.
\end{enumerate}

\paragraph{Properties of the $p^\best$ algorithm:}
Let $p$ be any extended chronological (incremental) policy of
length $l(p)\!\leq\!\tilde l$ and computation time per cycle
$t(p)\!\leq\!\tilde t$, for which there exists a proof of VA($p$)
of length $\leq\!l_P$. The algorithm $p^\best$, depending on
$\tilde l$, $\tilde t$ and $l_P$ but not on $p$, has always higher
rating than any such $p$. The setup time of $p^\best$ is
$t_{setup}(p^\best)\!=\!O(l_P^2\!\cdot\!2^{l_P})$ and the
computation time per cycle is $t_{cycle}(p^\best)\!=\!O(2^{\tilde
l}\!\cdot\!\tilde t)$. Furthermore, for $\tilde t,\tilde
l\!\to\!\infty$, $p^\best$ converges to the behavior of the AI$\xi$
model.

Roughly speaking, this means that if there exists a computable
solution to some AI problem at all, then the explicitly
constructed algorithm $p^\best$ is such a solution. Although this
claim is quite general, there are some limitations and open
questions, regarding the setup time regarding the necessity that
the policies must rate their own output, regarding true but not
efficiently provable VA($p$), and regarding ``inconsistent''
policies \cite{Hutter:00f}.

\section{Outlook \& Discussion}\label{secOutlook}
This section contains some discussion and remarks on otherwise
unmentioned topics.

\paragraph{Value bounds:}
Rigorous proofs of value bounds for the AI$\xi$ theory are the
major theoretical challenge -- general ones as well as tighter
bounds for special environments $\mu$. Of special importance are
suitable (and acceptable) conditions to $\mu$, under which $\hh
y_k$ and finite value bounds exist for infinite $Y$, $X$ and $m$.

\paragraph{Scaling AI$\xi$ down:}
\cite{Hutter:00f} shows for several examples how to integrate
problem classes into the AI$\xi$ model. Conversely, one can
downscale the AI$\xi$ model by using more restricted forms of
$\xi$. This could be done in a similar way as the theory of
universal induction has been downscaled with many insights to the
Minimum Description Length principle \cite{Li:92b,Rissanen:89} or
to the domain of finite automata \cite{Feder:92}. The AI$\xi$
model might similarly serve as a super model or as the very
definition of (universal unbiased) intelligence, from which
specialized models could be derived.

\paragraph{Applications:}
\cite{Hutter:00f} shows how a number of AI problem classes,
including {\em sequence prediction}, {\em strategic games}, {\em
function minimization} and {\em supervised learning} fit into
the general AI$\xi$ model. All problems are claimed to be formally
solved by the AI$\xi$ model. The solution is, however, only
formal, because the AI$\xi$ model is uncomputable or, at best,
approximable. First, each problem class is formulated in its
natural way (when $\mu^{\mbox{\tiny problem}}$ is known) and then
a formulation within the AI$\mu$ model is constructed and their
equivalence is proven. Then, the consequences of replacing $\mu$
by $\xi$ are considered. The main goal is to understand
how the problems are solved by AI$\xi$. For more details see
\cite{Hutter:00f}.

\paragraph{Implementation and approximation:}
The AI$\xi^{\tilde t\tilde l}$ model suffers from the same large
factor $2^{\tilde l}$ in computation time as Levin search for
inversion problems
\ifijcai\cite{Levin:73}.
\else\cite{Levin:73,Levin:84}.
\fi
Nevertheless, Levin
search has been implemented and successfully applied to a variety
of problems \cite{Schmidhuber:97nn,Schmidhuber:97bias}. Hence, a direct
implementation of the AI$\xi^{\tilde t\tilde l}$ model may also be
successful, at least in toy environments, e.g.\ prisoner problems.
The AI$\xi^{\tilde t\tilde l}$ algorithm should be regarded only
as the first step toward a {\em computable universal AI model}.
Elimination of the factor $2^{\tilde l}$ without giving up
universality will probably be a very difficult task. One could try
to select programs $p$ and prove VA($p$) in a more clever way than
by mere enumeration. All kinds of ideas like, heuristic search,
genetic algorithms, advanced theorem provers, and many more could
be incorporated. But now we have a problem.

\paragraph{Computability:}
We seem to have transferred the AI problem just to a different
level. This shift has some advantages (and also some
disadvantages) but presents, in no way, a solution. Nevertheless,
we want to stress that we have reduced the AI problem to (mere)
computational questions. Even the most general other systems the
author is aware of, depend on some (more than complexity)
assumptions about the environment, or it is far from clear whether
they are, indeed, universally optimal. Although computational
questions are themselves highly complicated, this reduction is a
non-trivial result. A formal theory of something, even if not
computable, is often a great step toward solving a problem and has
also merits of its own (see previous paragraphs).

\paragraph{Elegance:}
Many researchers in AI believe that intelligence is something
complicated and cannot be condensed into a few formulas. They
believe it is more a combining of enough {\em methods} and much
explicit {\em knowledge} in the right way. From a theoretical
point of view, we disagree as the AI$\xi$ model is simple and
seems to serve all needs. From a practical point of view we agree
to the following extent. To reduce the computational burden one
should provide special purpose algorithms ({\em methods}) from the
very beginning, probably many of them related to reduce the
complexity of the input and output spaces $X$ and $Y$ by
appropriate pre/post-processing methods.

\paragraph{Extra knowledge:}
There is no need to incorporate extra {\em knowledge} from the
very beginning. It can be presented in the first few cycles in
{\it any} format. As long as the algorithm that interprets the
data is of size $O(1)$, the AI$\xi$ system will 'understand' the
data after a few cycles (see \cite{Hutter:00f}). If the
environment $\mu$ is complicated but extra knowledge $z$ makes
$K(\mu|z)$ small, one can show that the bound (\ref{eukdistxi})
reduces to $\ln 2\!\cdot\!K(\mu|z)$ when $x_1\!\equiv\!z$, i.e.\
when $z$ is presented in the first cycle. Special purpose
algorithms could also be presented in $x_1$, but it would be
cheating to say that no special purpose algorithms have been
implemented in AI$\xi$. The boundary between implementation and
training is blurred in the AI$\xi$ model.

\paragraph{Training:}
We have not said much about the training process itself, as it is
not specific to the AI$\xi$ model and has been discussed in
literature in various forms and disciplines. A serious discussion
would be out of place. To repeat a truism, it is, of course,
important to present enough knowledge $x'_k$ and evaluate the
system output $y_k$ with $r_k$ in a reasonable way. To maximize
the information content in the reward, one should start with
simple tasks and give positive reward to approximately
the better half of the outputs $y_k$, for instance.

\paragraph{The big questions:}
\cite{Hutter:00f} contains a discussion of the ``big'' questions
concerning the mere existence of any computable, fast, and elegant
universal theory of intelligence, related to non-computable $\mu$
\cite{Penrose:94} and the `number of wisdom' $\Omega$
\cite{Chaitin:75,Chaitin:91}.


\end{document}